\documentclass[11pt]{article}

\usepackage[final]{acl}

\usepackage{times}
\usepackage{latexsym}
\usepackage{amsmath}
\usepackage{subcaption}
\usepackage{xcolor}

\usepackage[T1]{fontenc}

\usepackage[utf8]{inputenc}

\usepackage{microtype}

\usepackage{inconsolata}

\usepackage{graphicx}

\usepackage{float}

\title{Can We Do Interpretable NLI with Graphs Based on Atomic Propositions?}

\author{
  \textbf{Younes Boufouss}\textsuperscript{1}\begin{NoHyper}\thanks{Equal contribution.}\end{NoHyper}, \textbf{Luc Pommeret}\textsuperscript{1,2}\footnotemark[1] \\
  \textbf{Thomas Gerald}\textsuperscript{1}, \textbf{Patrick Paroubek}\textsuperscript{1}, \textbf{Sophie Rosset}\textsuperscript{1} \\
  \textsuperscript{1}Université Paris-Saclay, CNRS, LISN, Orsay, France \\
  \textsuperscript{2}SCIAM \\
  \small{
    \textbf{Correspondence:} \href{mailto:younes.boufouss@universite-paris-saclay.fr}{younes.boufouss@universite-paris-saclay.fr}
  }
}

\begin{document}
\maketitle
\begin{abstract}
While Large Language Model (LLM)-based Natural Language Inference (NLI) systems achieve high accuracy, their decision-making processes lack auditable structures. This paper explores whether NLI can be performed using only interpretable, graph-based representations of evidence. We introduce a fully graph-based pipeline where the classifier never directly processes the input text. Instead, sentences are decomposed into atomic propositions, converted into \texttt{ConceptNet} triples via constrained decoding, and represented as three graphs per pair: premise, hypothesis, and a retrieved \texttt{ConceptNet} subgraph. These graphs are then fed into a fine-tuned 0.8-billion-parameter language model.
On the SNLI dataset, our pipeline achieves 89.7\% accuracy, just 1.9 points below an identically trained text-based model. On ANLI, it matches the published performance of \texttt{RoBERTa-large} on rounds R2 and R3 (48.0\% vs.\ 48.9\% and 44.9\% vs.\ 44.4\%) but trails by 16 points on R1, resulting in an overall gap of 9 to 14 points compared to its text counterpart. We term this gap the \emph{price of interpretability} and demonstrate that it stems from representational limitations rather than data constraints. Ablation studies further reveal that graphs and text are complementary: combining both modalities achieves 92.1\% accuracy on SNLI.
\end{abstract}

\section{Introduction}

Natural Language Inference (NLI) is a fundamental task in Natural Language Understanding (NLU). Accurate predictions in NLI require multi-level reasoning, encompassing lexical (e.g., synonyms, antonyms), syntactic (e.g., negation, quantifiers), and semantic dimensions (e.g., logical entailment, common-sense knowledge). While current systems achieve state-of-the-art performance on benchmarks like SNLI~\citep{bowman-etal-2015-large} and MultiNLI~\citep{williams-etal-2018-broad}, these datasets are not without flaws. For instance, annotation biases, negation, and vagueness are strongly correlated with specific classes, enabling even shallow models to perform surprisingly well. \citet{gururangan-etal-2018-annotation} demonstrated that a simple text classifier trained solely on the hypothesis achieves 67\% accuracy on SNLI and 53\% on MultiNLI. This suggests that a significant portion of the high performance in these benchmarks may stem from exploiting linguistic artifacts rather than genuine reasoning capabilities.

\begin{figure*}[!t]
\centering
\includegraphics[width=\textwidth]{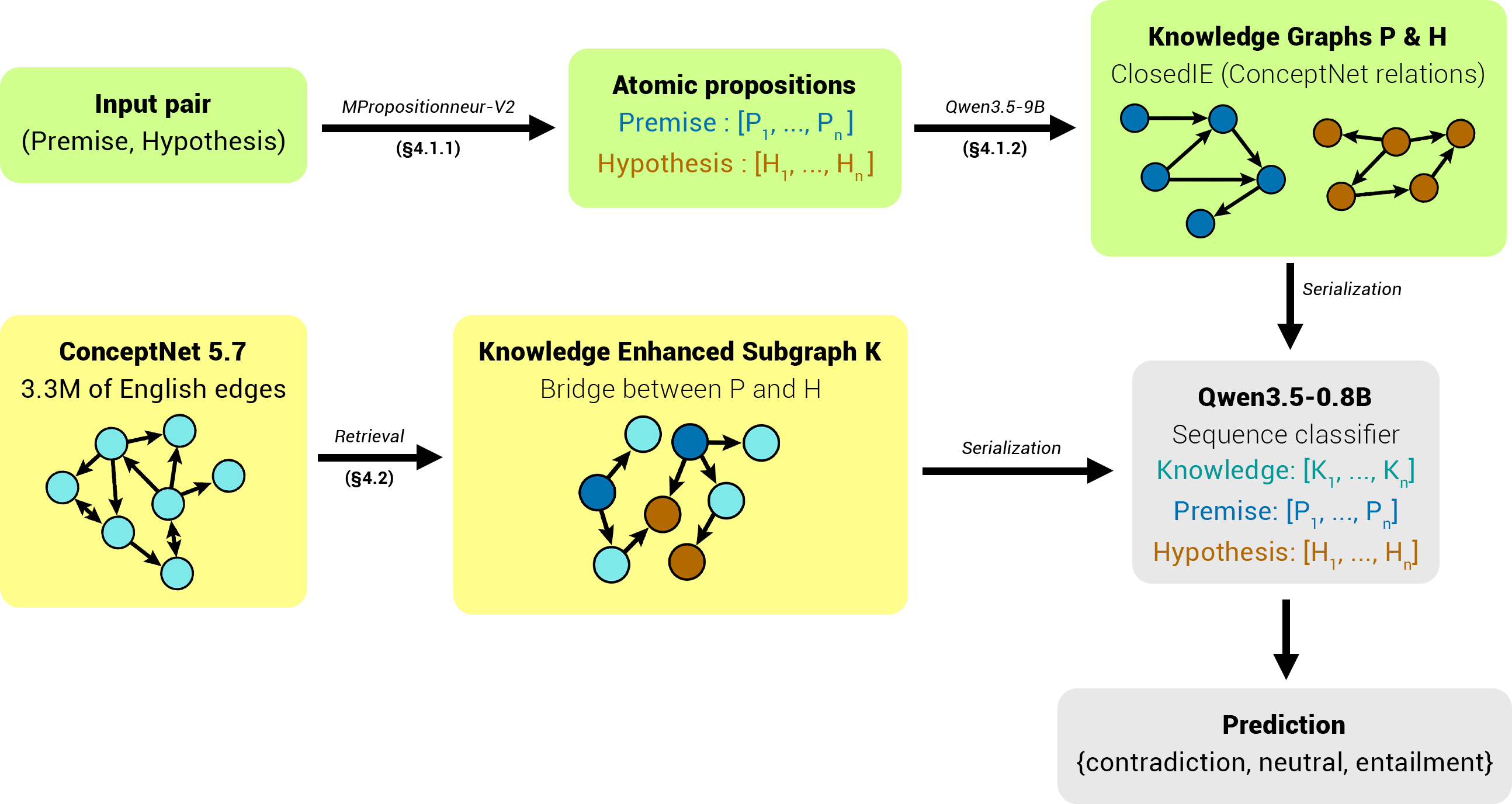}
\caption{The four stages of the pipeline. Both sentences of the pair are decomposed
into atomic propositions, each proposition is converted into \texttt{ConceptNet} triples under
a constrained JSON schema to form the premise graph $P$ and the hypothesis graph $H$,
a third graph $K$ is retrieved from \texttt{ConceptNet}, and the three graphs are serialised in
the order $K, P, H$ for the classifier. No natural language text reaches the classification stage.}
\label{schema}
\end{figure*}

To address these artifacts, ANLI~\citep{anli} was introduced as a more robust benchmark. Constructed through an iterative human-in-the-loop protocol, ANLI consists of three rounds (R1, R2, and R3) of increasing difficulty. In each round, annotators are tasked with crafting hypotheses that state-of-the-art models would misclassify, given a premise. These examples are then verified by other annotators to ensure their validity. After each round, the adversarial examples are incorporated into the training data, and the model is retrained.

Given this design, ANLI provides a natural testbed to evaluate the effectiveness of our knowledge graph-based approach. By assessing performance on ANLI, we can determine whether our system leverages structured knowledge to achieve genuine reasoning or if it, too, falls back on superficial patterns.

In this context, two successive measures of the same metric --- Accuracy --- are typically used. The first measure is obtained by providing raw text to a large pretrained model, which contains no explicit reasoning cues. Interpretability, however, is achieved by constraining the decision-making process to rely on interpretable intermediate structures (\textit{e.g.}, knowledge graphs).

In this work, we quantify the \emph{price of interpretability}, \textit{i.e.}, the performance gap between these two approaches.

Our central research question is: \emph{Do graphs suffice?} In other words, can structured representations alone achieve competitive performance, or is there an inherent trade-off between interpretability and accuracy? Our goal is to measure and analyse this trade-off systematically.

\paragraph{Our contribution.}

Here, we provide a pipeline to perform the NLI task entirely on knowledge graphs, in the syntax of \texttt{ConceptNet}~\citep{conceptnet}. The graphs are extracted from atomic propositions as in~\citet{pommeret2026llmbasedatomicpropositionshelp}, augmented with a retrieved \texttt{ConceptNet} subgraph, serialised, and classified by a fine-tuned \texttt{Qwen3.5-0.8B-Base}. The design guarantees that no raw text reaches the classifier, so every piece of information used for the decision is present in an interpretable list of triplets.\footnote{Code: \url{https://github.com/l-pommeret/graph-nli}; see Appendix~\ref{repro}.} Our contributions are as follows:
\begin{enumerate}
    \item A full graph-based NLI pipeline, featuring a 28-relation extraction vocabulary derived from \texttt{ConceptNet}\footnote{We use ConceptNet 5.7. The 28 relations are those selected for our extraction schema (Appendix~\ref{prompt}); ConceptNet contains additional relations.}, and a constrained JSON schema ensuring structural consistency (see Section~\ref{extract}).
    \item Quantification of the \emph{price of interpretability}: we measure the accuracy drop between text-based and graph-based models \textit{ceteris paribus}: $-1.9$ points on SNLI and $-9$ to $-14$ points on ANLI (see Table~\ref{tab:prix-interpretabilite}).
    \item Representation, not data, as the bottleneck: the performance gap stems from structural limitations rather than data scarcity. Adding 600,000 training pairs from MNLI and FEVER-NLI yields only $+0.6$ points on ANLI (see Section~\ref{scaling}).
    \item Ablation studies to isolate key factors: (i) the contribution of external knowledge from \texttt{ConceptNet}, and (ii) the combination of graph and text models (see Section~\ref{ablations}).
\end{enumerate}

\section{Related Work}
Leveraging external knowledge for Natural Language Understanding (NLU) tasks is a well-established approach. In Question Answering (QA), systems like QA-GNN~\citep{yasunaga-etal-2021-qa} and KagNet~\citep{lin-etal-2019-kagnet} bridge the semantic gap between questions and candidate answers by integrating \texttt{ConceptNet}~\citep{conceptnet} as an external knowledge base. For Natural Language Inference (NLI), KIM~\citep{Chen-Qian:2018:ACL} enhances lexical understanding by incorporating WordNet-based relations, such as synonymy, antonymy, hyperonymy, and hyponymy. Similarly, KGNLI~\citep{wang-etal-2020-knowledge} extracts key concepts from premises and hypotheses to construct a knowledge subgraph linking the two.

While decomposing premises and hypotheses into atomic propositions improves interpretability and enables the diagnosis of logical flaws, this approach alone does not inherently boost accuracy without fine-tuning the models~\citep{srikanth2025nlimicroscopeatomichypothesis, huang2026atomicsnlifinegrainednaturallanguage}. The pipeline described in Section~\ref{sec:approach} addresses this limitation by making the atomic propositions the input of the triplet extraction stage.

Closest to our setting are the verification systems whose verdict is a function of an explicit structure rather than of the sentences themselves. ProoFVer~\citep{krishna-etal-2022-proofver} generates a natural-logic proof --- a sequence of lexical mutations between spans of the claim and of the retrieved evidence, each marked with a natural-logic operator --- and reads the verdict off the operator sequence alone, so that the explanation is faithful by construction. \citet{yuan-vlachos-2024-zero} decompose claim and evidence into semantic triples, augment them with an external knowledge graph, and pass the result to an off-the-shelf NLI model, in a zero-shot setting. We share with both previous works the commitment to a decision that depends on the structure alone, and we differ on three points: the decomposition goes through atomic propositions before the triplet extraction, the relation inventory is a closed enumeration of 28 \texttt{ConceptNet} relations enforced by constrained decoding rather than an open predicate set, and the classifier is fine-tuned on the serialised graphs, which is what makes the \textit{ceteris paribus} comparison with a text model of the same backbone possible.

Finally, our classifier processes graphs as serialised text. Serialising structured inputs for language models rather than encoding them with a dedicated graph network is now a common choice~\citep{fatemi-etal-2024}, and we follow it: it leverages the pretrained semantics of relation names and ensures that each triplet remains individually accessible to the model's attention mechanism. While this approach offers clear advantages, a systematic comparison with graph encoders remains an avenue for future work.

\section{Proposed Approach}
\label{sec:approach}

Atomic facts are first extracted from both the premise and the hypothesis, and only then converted into \textit{(subject, relation, object)} triplets. This decomposition improves triplet extraction~\citep{pommeret2026llmbasedatomicpropositionshelp} by simplifying the task assigned to the extractor: rather than isolating multiple facts from a complex sentence in a single pass, it processes one minimal semantic unit at a time, which reduces the number of facts to extract per call and ensures that no atomic fact from the source sentence is omitted.

The pipeline maps a pair (premise, hypothesis) to a label in the set \{entailment, neutral, contradiction\} through four stages, so that no original text from the pair takes part in the final decision (see Figure~\ref{schema}).

\begin{enumerate}
    \item \textbf{Atomisation}. Both elements from the pair are decomposed into atomic propositions by \texttt{MPropositionneur-V2-large}, a model distilled for multilingual atomisation with coreference resolution~\citep{pommeret2026llmbasedatomicpropositionshelp}.
    \item \textbf{Triplet extraction.} Each atomic proposition is converted into a set of (subject, relation, object) triplets by \texttt{Qwen3.5-9B}\footnote{\url{https://huggingface.co/Qwen/Qwen3.5-9B}} under a constrained JSON schema whose relation field is an enumeration of 28 \texttt{ConceptNet} relations. The union of the triplets of its propositions forms the graph $P$ of the premise, and the graph $H$ of the hypothesis.
    \item \textbf{Knowledge enhancement.} A third graph, $K$, is retrieved from the 3.3M edges of the \texttt{ConceptNet} base: bridge edges (1 and 2 hops) between the entities that occur in $P$ but not in $H$, and that occur in $H$ but not in $P$, plus a one-hop neighbourhood of each entity.
    \item \textbf{Classification.} The three graphs are serialised into a single string, in the following order: $K, P, H$, and fed to \texttt{Qwen3.5-0.8B-Base},\footnote{\url{https://huggingface.co/Qwen/Qwen3.5-0.8B-Base}} fine-tuned on the training sets of different NLI benchmarks.
\end{enumerate}

\paragraph{Why we do not give raw text to the LLM.} A model that reads both the text and the graph will use the graph only where the text is insufficient. Thus, any interpretation of its graph usage is \textit{post-hoc}. By removing the text, we make the intermediate structure decisive in the sense that \textit{an error in the graph is an error in the prediction}. Conversely, every prediction can be traced to a finite list of triplets. The cost of the interpretability choice is presented in Table~\ref{tab:prix-interpretabilite}.

\section{Experimental Protocol}

This section follows the pipeline in the order in which it is applied. Section~\ref{sec:graphconstruction} covers graph construction, first the atomisation of the sentences into propositions (Section~\ref{sec:propositionneur}), then the extraction of triplets from each proposition (Section~\ref{extract}); Section~\ref{sec:knowledge} describes the retrieval of the external subgraph $K$; and Section~\ref{sec:serialisation} the serialisation of the three graphs and the fine-tuning of the classifier. Each stage constrains the next: the propositions bound what the extractor can see, and the extracted entities bound what can be retrieved from \texttt{ConceptNet}.

\subsection{Graph Construction}\label{sec:graphconstruction}

\subsubsection{Propositionneur}\label{sec:propositionneur}

Sentences are atomised by \texttt{MPropositionneur-V2-large}, a fine-tuning of \texttt{Qwen3}, distilled from a larger teacher on Wikipedia chunks in six European languages~\citep{pommeret2026llmbasedatomicpropositionshelp}. The model is prompted with \texttt{Atomize: \{sentence\}} and its output is constrained to a JSON array of strings under \texttt{vLLM}~\citep{Kwon-etal-2023}. Decoding is greedy (temperature $T=0$).

Long adversarial premises from ANLI make the propositionneur degenerate into loops of repetition in 7\% of ANLI sentences. We therefore apply a cleaning step: case-insensitive and whitespace-insensitive deduplication, with removal of any proposition longer than 1.6 times the source sentence plus twenty characters, and a cap of 16 propositions per sentence, with a fallback to the source sentence when the filter empties the list. SNLI and ANLI (R1, R2 and R3) have been atomised.

\subsubsection{Information Extraction}\label{extract}

\paragraph{Extraction.} Extraction uses \texttt{Qwen3.5-9B} served by \texttt{vLLM}, with few-shot prompting: the prompt provides the list of relations with their descriptions, a set of encoding rules, and nine annotated examples (see Appendix~\ref{prompt}). Importantly, decoding is constrained by a JSON schema in which the relation field is an \texttt{enum} over the 28 relations, so out-of-vocabulary relations are impossible by design. The schema also limits the output to 8 triplets: the prompt asks for at least two, while the schema enforces at least one.

\paragraph{Relation vocabulary.} The extraction vocabulary contains 28 relations, all of them \texttt{ConceptNet} relations~\citep{conceptnet}. Using only \texttt{ConceptNet} relation names has benefits: the extracted graphs are directly alignable with the external base used at stage 3, and the relation names carry pretrained semantics that the classifier (fine-tuned LLM) can exploit.

\paragraph{Prompt rules.} The extractions produce graphs that are locally plausible, but globally problematic, because the same fact is encoded differently in $P$ and $H$, and the two graphs fail to align it. Therefore, the prompt fixes a normal form, which is enforced by the following rules:

\begin{enumerate}
    \item \textbf{The nodes are lemmas.} A node is a lower-case English lemma reduced to its head word: modifiers are represented separately (\textit{e.g.}, \texttt{old\_man} is forbidden, and \texttt{man} plus \texttt{[man, HasProperty, old]} is required). Multi-word nodes are reserved for proper names and lexical compounds. Non-intersective adjectives (\textit{e.g.}, \texttt{former\_senator}, \texttt{fake\_gun}) are explicit exceptions, since detaching them would change the truth conditions. Verbs and their objects are represented separately, except when the negation rule below requires a full verb phrase.

    \item \textbf{Argument structure is fixed.} The agent of a verb always takes \texttt{CapableOf} and the patient always \texttt{ReceivesAction}; both are from a transitive verb. The passive and active variants are normalised to a unique form, so that \textit{the book was written} and \textit{someone wrote the book} yield the same triplet.

    \item \textbf{Adjuncts are typed.} Beneficiaries and recipients use \texttt{HasContext}, destinations \texttt{MotivatedByGoal}, locations \texttt{AtLocation}, instruments \texttt{UsedFor}, and temporal expressions \texttt{HasContext}.

    \item \textbf{Cardinality is an object.} For example \textit{two men} is expressed as \texttt{[man, HasProperty, two]}, but a bare plural expresses no numeral.

    \item \textbf{Negation is a relation.} Negated facts use the Not* relations with the full verb phrase as a tail. This is what allows contradiction to be visible in the graph.

    \item \textbf{Created works have a fixed direction.} The created work is always the subject of \texttt{CreatedBy}.
\end{enumerate}

Every content word of the proposition source must give (at least) one triple. After the decoding, the nodes are normalised to the \texttt{ConceptNet} form with the \texttt{spaCy} lemmatiser.\footnote{\url{https://spacy.io/api/lemmatizer}} Function words are dropped (except \texttt{not}, \texttt{no} and \texttt{never}). The remaining tokens are lemmatised, lower-cased and joined by underscores.

\paragraph{Prompt validation.} The rules described above were not written \textit{a priori}. They are the result of two corrective iterations driven by an LLM-as-judge panel (with three evaluation lenses over 36 items). That evaluation surfaced five systematic failures in the first prompt: hallucinated cardinality on bare plurals, an agent/patient inversion, the beneficiaries encoded as the patients, worn or printed identifiers encoded as quantities, and an inversion of the \texttt{CreatedBy} direction.

\subsection{Knowledge Enhancement}\label{sec:knowledge}

The knowledge base is \texttt{ConceptNet 5.7}, restricted to English assertions, excluding the relations from DBpedia, and the relation \texttt{ExternalURL} (that is not used here). High-degree hubs are truncated to their 64 best neighbours, ranked by informativeness first and by weight second, so that a frequent node (\textit{e.g.}, \texttt{man}) does not invade every subgraph.

For a pair, let $\mathcal{P}$ and $\mathcal{H}$ be the sets of the entities that occur in $P$ and $H$, anchored to \texttt{ConceptNet} by exact match, or, if it does not suffice, by lexical head (\textit{e.g.}, \texttt{winter\_hat} becomes \texttt{hat}). The subgraph $K$ is built in three passes:

\begin{enumerate}
    \item \textbf{Direct bridges.} Edges linking an entity of $\mathcal{P}\setminus \mathcal{H}$ (\textit{i.e.}, entities of $\mathcal{P}$ that are not in $\mathcal{H}$), to an entity of $\mathcal{H}\setminus\mathcal{P}$, ranked by weight. Shared entities are excluded with a reason: they are already aligned by string identity, so a bridge between them carries no information.

    \item \textbf{Two-hop bridges.} Paths $p \rightarrow m \rightarrow h$ with $p \in \mathcal{P} \setminus \mathcal{H}, h \in \mathcal{H} \setminus \mathcal{P}$ and the pivot $m$ outside $\mathcal{P} \cup \mathcal{H}$, ranked by the weight of their weakest edge. This pass is intended to capture transitive evidence such as \texttt{london PartOf england PartOf europe}, which no single triplet of $P$ or $H$ contains.

    \item \textbf{Informative neighbourhood.} Up to 5 one-hop edges per entity, and excluding lexical relations (as \texttt{RelatedTo, FormOf, DerivedFrom, Synonym, SimilarTo}). Lexical relations are admitted as bridges, where they are genuinely informative, but not as background, where they only add noise.
\end{enumerate}

Bridges are capped at 30 direct edges and 60 two-hop edges, the neighbourhood at 5 edges per entity, and $K$ at 80 edges in total. Note that $K$ may contain eight relations the extractor cannot produce (\texttt{RelatedTo}, \texttt{Entails}, \texttt{FormOf}, \texttt{DerivedFrom}, \texttt{ObstructedBy}, \texttt{HasFirstSubevent}, \texttt{HasLastSubevent}, \texttt{SymbolOf}), so the serialised input uses 36 relation types in total.

\subsection{Serialisation and Fine-tuning of Qwen}\label{sec:serialisation}

Each pair is rendered as three labelled sections, with the triplets separated by semicolons:

\begin{verbatim}
    knowledge: london PartOf england ; ...
    premise: woman HasProperty two ; ...
    hypothesis: sister CapableOf hug ; ... 
\end{verbatim}

This ordering places external knowledge $K$ first, followed by the premise graph $P$, and finally the hypothesis graph $H$, closest to the classification head. Truncation is on the left, so an input that exceeds the length limit loses external knowledge before premise information, and premise information before the hypothesis.

\paragraph{What an auditor sees.} Figure~\ref{fig:worked} follows one pair through the four stages. It is the whole of what the classifier receives, and therefore the whole of what has to be checked to accept or reject the prediction. Auditing it is a finite task with three questions, each answerable without running the model. Does every triplet of $P$ and $H$ follow from its atomic proposition? Does every atomic proposition follow from its sentence? Is the bridge in $K$ that carries the decision --- here \texttt{girl IsA child} and \texttt{park AtLocation outdoors}, without which nothing in $P$ meets \texttt{child} or \texttt{outdoors} --- a licit inference step? A wrong prediction is thus localised to a specific line rather than attributed to the model as a whole, and a corrected line can be fed back through the classifier unchanged. This is the sense in which we use \emph{interpretable}: the object of the audit is the input, not a saliency map over it.

\floatstyle{boxed}
\restylefloat{figure}
\begin{figure}[t]
\small
\setlength{\fboxsep}{4pt}
\noindent\textbf{Pair.} \textit{premise:} Two young girls are playing soccer in a park while their coach watches. --- \textit{hypothesis:} The children are outdoors. --- \textit{gold:} entailment.

\medskip
\noindent\textbf{1. Atomic propositions.}
\begin{itemize}\setlength\itemsep{0pt}\setlength\parskip{0pt}
  \item[$P_1$] Two young girls are playing soccer.
  \item[$P_2$] The girls are in a park.
  \item[$P_3$] The coach of the girls watches the girls.
  \item[$H_1$] The children are outdoors.
\end{itemize}

\medskip
\noindent\textbf{2--3. Serialised input to the classifier.}
\begin{quote}
\footnotesize\ttfamily
knowledge: girl IsA child ; park AtLocation outdoors ; soccer AtLocation park\\
premise: girl HasProperty two ; girl HasProperty young ; girl CapableOf play ; soccer ReceivesAction play ; play AtLocation park ; coach CapableOf watch ; girl ReceivesAction watch ; girl HasA coach\\
hypothesis: child AtLocation outdoors
\end{quote}

\medskip
\noindent\textbf{4. Prediction.} entailment.
\caption{A pair carried through the pipeline. No token of the two sentences reaches stage 4: the decision rests on the twelve triplets above. The entailment is not derivable from $P$ and $H$ alone --- \texttt{child} and \texttt{outdoors} occur nowhere in $P$ --- and is licensed by the two bridge edges retrieved in $K$.}
\label{fig:worked}
\end{figure}

The classifier is \texttt{Qwen3.5-0.8B-Base}, with a three-way sequence classification head. Training uses a batch size of 8, with 8 gradient accumulation steps (so an effective batch size of 64), the learning rate is 2e-5, with cosine scheduler and a 3\% warmup. For regularisation, we use a weight decay of 0.01, on 2 epochs. Training is in \texttt{bf16}, with a maximum sequence length of 1024 tokens (1536 on ANLI, whose premises are longer).

\paragraph{Checkpoint selection.} The checkpoint is selected on the \textit{validation} split of the dataset, not on the \textit{test}. All figures below have been obtained following this protocol.

\section{Datasets and Benchmarks}

The two evaluation benchmarks are chosen because they sit at the two ends of the range over which a structural bottleneck can be measured. SNLI~\citep{bowman-etal-2015-large} is the favourable end: its premises are single-sentence image captions describing one visible event, with a short concrete vocabulary, few named entities and almost no temporal or modal content, so the normal form of Section~\ref{extract} has a faithful target for nearly every content word. ANLI~\citep{anli} is the adversarial end: its premises are multi-sentence passages, drawn from Wikipedia for R1 and R2 and from a wider range of genres for R3, and its hypotheses were written by annotators with the explicit goal of defeating a strong text model, so they turn on exactly the fine-grained detail --- a date, a number, a comparison, an embedded clause --- that a fixed triplet vocabulary is likely to compress away. Holding the pipeline and the backbone fixed across the two, the gap moves from 1.9 to 14 points as the input moves from captions to adversarial passages, which is what isolates the cost of the representation rather than of the model.

For the data-scaling experiment of Section~\ref{scaling} only, we additionally extract MNLI~\citep{williams-etal-2018-broad}, which adds ten written and spoken genres, and FEVER-NLI~\citep{thorne-etal-2018, nie-etal-2019-combining}, which adds retrieved encyclopaedic evidence, using the evidence as the premise. Together they contribute about 600,000 further training pairs. They are used as additional training data, never as test sets.

The full SNLI training set (549,367 pairs) is processed, yielding 889,944 unique atomic propositions. ANLI R1 to R3 yield 296,585 atomic propositions. MNLI has 953,734 unique propositions, and FEVER-NLI 507,593. Extraction produces an average of 3.6 triplets per proposition.

\subsection{Metrics}

We report accuracy. ANLI test sets contain 1000 to 1200 examples (depending on the round).

\subsection{Defining the Price of Interpretability}

We define as \textit{price of interpretability} the drop between the original fine-tuning of \texttt{Qwen3.5-0.8B} on the texts of premise and hypothesis (with the format \texttt{premise: ...\textbackslash nhypothesis: ...}) and the \textit{ceteris paribus} fine-tuning of \texttt{Qwen3.5-0.8B} only on graphs extracted as we show in Section~\ref{extract}. We use the same backbone, optimiser and nominal number of epochs (see Appendix~\ref{repro}); differences in seed coverage and completed training budgets are discussed in the \hyperref[limitations]{Limitations section}. The input is raw premise and hypothesis text for the text model, and serialised graphs for the graph model. The quantity is therefore not a comparison against the state-of-the-art, but a measurement of what the graph-interpretability costs. Since the text baseline is a single run while the graph model has three seeds, the SNLI price is defined against the three-seed mean.

\section{Results and Analysis}

\subsection{Main Results}

Table~\ref{tab:snli} reports SNLI. The graph pipeline reaches $0.897 \pm 0.006$ accuracy over three seeds. Below it, the zero-shot control matters for interpreting all of this. Without fine-tuning, the 0.8B base model scored by label likelihood is at chance on every split (0.343 on SNLI, 0.334 to 0.335 on ANLI). The zero-shot score is the same whether the input is text or graphs. All the competence reported below comes from fine-tuning, and none of it from the backbone model.

Table~\ref{tab:anli} reports ANLI. Transferring the SNLI model zero-shot gives 0.285 for R1, 0.330 for R2 and 0.328 for R3, that is at, or below, chance, which is expected because ANLI is designed to produce examples that fool models which are good at SNLI. After a second training phase, on ANLI R1 to R3, starting from the SNLI checkpoint with results shown in Table~\ref{tab:snli}, the graph model reaches 0.576, 0.480 and 0.449, \textit{i.e.}, +29.1 points on R1 over the zero-shot transfer, with a graph-only input.
Sequential fine-tuning costs SNLI retention (dropping from 0.892 to 0.782).

Two readings of Table~\ref{tab:anli} pull in two directions: against the literature, the graph-only model is within one point of the published \texttt{RoBERTa-large}~\citep{liu2019roberta}\footnote{\url{https://huggingface.co/FacebookAI/roberta-large}} test accuracy on R2 (0.480 vs.\ 0.489) and slightly above it on R3 (0.449 vs.\ 0.444), while remaining 16 points behind on R1 (0.576 vs.\ 0.738)~\citep{anli}; it sits at the level of \texttt{BERT-large}~\citep{devlin-etal-2019-bert} on all three rounds (0.574, 0.483 and 0.435). An interpretable-by-design system that never sees the text is therefore competitive with strong text encoders on the two hardest rounds. Against the \textit{ceteris paribus} control, the same graphs lose 9 to 14 points to their own text counterpart. The interpretability gap only means something when measured against the same backbone, so this is what we call the price of interpretability.

\begin{table}[t]
\centering
\small
\setlength{\tabcolsep}{3.5pt}
\begin{tabular}{@{}llc@{}}
\hline
System & Input & SNLI \\
\hline
Zero-shot 0.8B-Base & text & 0.343 \\
Zero-shot 0.8B-Base & graphs & 0.343 \\
RoBERTa-large (ours) & text & 0.899 \\
\hline
Qwen3.5-0.8B (ours) & graphs & $0.897 \pm 0.006$ \\
Qwen3.5-0.8B & text & 0.916 \\
Qwen3.5-0.8B & text+graphs & \textbf{0.921} \\
\hline
\end{tabular}
\caption{SNLI test accuracy. Our graph model is the mean over three seeds
(0.892, 0.894 and 0.904). All other rows are seed 0.}
\label{tab:snli}
\end{table}

\begin{table}[t]
\centering
\small
\setlength{\tabcolsep}{2.5pt}
\begin{tabular}{@{}lcccc@{}}
\hline
Model & R1 & R2 & R3 & SNLI \\
\hline
\multicolumn{5}{@{}l}{\textit{Graphs only}} \\
Zero-shot transfer & 0.285 & 0.330 & 0.328 & 0.892 \\
Train on ANLI only & \textbf{0.576} & \textbf{0.480} & 0.449 & 0.782 \\
ANLI + SNLI replay & 0.565 & 0.462 & \textbf{0.463} & \textbf{0.887} \\
\hline
\multicolumn{5}{@{}l}{\textit{Text}} \\
RoBERTa-large (ours) & 0.580 & 0.359 & 0.357 & 0.899 \\
RoBERTa-large (published) & 0.738 & 0.489 & 0.444 & -- \\
BERT-large (published) & 0.574 & 0.483 & 0.435 & -- \\
Qwen3.5-0.8B & 0.704 & 0.572 & 0.553 & 0.910 \\
\hline
\end{tabular}
\caption{ANLI accuracy, with SNLI transfer in the last column. Bold marks the best
graph-only system per column. The published \texttt{RoBERTa-large} and \texttt{BERT-large} results are
those of~\citet{anli}.}
\label{tab:anli}
\end{table}

\subsection{The Price of Interpretability}
\label{sec:price}

Table~\ref{tab:prix-interpretabilite} quantifies the trade-off. The price is small on SNLI (we see that the graph pipeline retains 97.9\% of the text model's accuracy), and big on ANLI, where it retains only 80\% on R1. The adversarial signal lives in what the structure of triplet extraction compresses away.

By inspecting the disagreements, we see four recurring categories accounting for most of the gap. Numerical reasoning (or order reasoning) survives only when the numeral is explicit. The prompt expressly forbids inventing one from a bare plural. Temporal reasoning is flattened: dates become undifferentiated with the \texttt{HasContext} relation, so \textit{since 1953} and \textit{in 1953} are indistinguishable (see the final paragraph of the \hyperref[limitations]{Limitations section}). Also, the coreferences on long premises are resolved by the propositionneur, and in case of error, it propagates through the pipeline. Finally, ANLI premises are long, and the fine-grained detail on which an adversarial hypothesis hinges is often the modifier that the normal form strips or the clause that the 8-triple cap drops.

\subsection{Ablations}
\label{ablations}

Table~\ref{tab:ablations} reports two ablations.

\paragraph{External knowledge is neutral on SNLI.} We see that removing $K$ (that is, the graph coming from \texttt{ConceptNet}) costs nothing (0.895 against 0.892). We can take it as a negative result: SNLI inference is so simple that the premise and hypothesis graphs suffice, and the retrieved \texttt{ConceptNet} subgraph is (at best) redundant. It does not follow that $K$ is useless in general. But on SNLI our multi-hop retrieval carries nothing.

\paragraph{Graphs and text are complementary.} Adding the graphs to the text raises the accuracy from 0.916 to 0.921. As a \textit{complement}, the graphs contribute.

\subsection{Is the Gap a Data Problem?}
\label{scaling}

A natural objection to Table~\ref{tab:prix-interpretabilite} is that the graph model is simply under-trained: graphs are a less familiar input format, so it may need more data to reach the same performance. We tested this directly. We extracted MNLI and FEVER-NLI with exactly the same pipeline, and ran a classical (in the literature) two-phase recipe: phase 1 on SNLI + MNLI + FEVER, and phase 2 on ANLI with SNLI replay. Phase 1 improves SNLI a little (0.896 against 0.892). Phase 2 on ANLI with SNLI replay then gives 0.568, 0.479 and 0.461 on R1 to R3, against 0.565, 0.462 and 0.463 without the additional data. We read this as evidence that the adversarial gap is representational, and not data-limited.

\begin{table}[t]
\centering
\small
\begin{tabular}{@{}lccc@{}}
\hline
 & Graph & Text & Price of Interp.\\
\hline
SNLI & 0.897 $\pm 0.006$ & \textbf{0.916} & $-1.9$ \\
ANLI R1 & 0.565 & \textbf{0.704} & $-13.9$ \\
ANLI R2 & 0.462 & \textbf{0.572} & $-11.0$ \\
ANLI R3 & 0.463 & \textbf{0.553} & $-9.0$ \\
\hline
\end{tabular}
\caption{Price of Interpretability: \texttt{Qwen3.5-0.8B} on raw text (\texttt{premise: ...\textbackslash nhypothesis: ...}) vs. the graph model, \textit{ceteris paribus}.}
\label{tab:prix-interpretabilite}
\end{table}

\begin{table}[t]
\centering
\small
\begin{tabular}{@{}lc@{}}
\hline
Configuration (SNLI test) & Accuracy \\
\hline
$P + H + K$ (full, graphs only) & $0.892$ \\
$P + H$, no external knowledge & 0.895 \\
\hline
Text only & 0.916 \\
Text + graphs & \textbf{0.921} \\
\hline
\end{tabular}
\caption{Ablations, all with the same \texttt{Qwen3.5-0.8B} backbone and protocol (seed 0).}
\label{tab:ablations}
\end{table}

\section{Conclusion}

The answer to our title question (can we do interpretable NLI with graphs based on atomic propositions?) is: yes on SNLI, and only partly on ANLI. We built an NLI system whose classifier never sees a natural-language sentence, and measured the performance cost of relying on graphs of atomic propositions alone. On SNLI, the cost is 1.9 points. On ANLI it is 9 to 14, depending on the ANLI round. The same system matches the published \texttt{RoBERTa-large} text model on ANLI R2 and R3 while taking no text as input.

Beyond that measurement, the pipeline itself is the contribution: atomic propositions are carried forward to the decision, and the combination of the \texttt{ConceptNet} vocabulary, the constrained extraction, and the serialised three-graph input makes every prediction traceable to a finite list of triplets. The cost of that guarantee is now measured, which is what a neuro-symbolic model needs if its intermediate representations are to be inspected by humans.

The practical consequence is that the \texttt{ConceptNet} triplet vocabulary, as we use it, is lossy in identifiable ways: numeral mentions, the temporal granularity, and the fine details of long premises. To close this gap, we therefore have to enrich the target structure, rather than scaling the classifier. Finally, since the interpretability we claim is the auditability of the intermediate structures (atomic propositions and triplets), it should be validated as such by a human study in which annotators repair a wrong prediction by editing the triplets. That would test the claim directly.
\section*{Limitations}
\label{limitations}

\paragraph{The interpretability is qualitative, not quantitative.} We do not measure the interpretability (\textit{i.e.}, the readability by humans) of atomic propositions and triplets of our pipeline explicitly. It remains to be tested in future work.

\paragraph{Validation of the graph by the evaluation.} We do not directly evaluate the quality of the graph, but evaluate it indirectly by means of NLI benchmarks. We want to conduct further evaluation on the quality of the graph, using other domains, or a direct approach.

\paragraph{Upstream errors are unrecoverable.} The guarantee that an error in the graph is an error in the prediction has a symmetric cost: the classifier has no text to fall back on, so any failure of the two upstream models propagates unchecked. Two are known and unmeasured. The atomiser degenerates into repetition loops on 7\% of ANLI sentences; our cleaning step removes the loops but is itself lossy, since the length filter, the cap of 16 propositions per sentence and the fallback to the source sentence can each drop a fact that the source did contain, and we do not measure how often they do. Coreference is resolved inside the atomiser, on long premises where it is hardest, and a wrong antecedent is silently frozen into the triplets of $P$. Quantifying the two --- by scoring propositions and triplets against a reference, rather than only through downstream accuracy --- is the most direct way to separate the price of the representation from the price of the extractors that build it.

\paragraph{Only English.} The propositionneur is multilingual, but the extraction vocabulary, the \texttt{ConceptNet} base, and every experiment, are in English. ConceptNet's coverage is culturally biased.

\paragraph{Single-seed baselines and unequal budgets.} The graph model is averaged over three seeds, but the text baselines and all ANLI models are single runs; two of the three graph seeds were also interrupted before the end of the second epoch. The SNLI price of interpretability should therefore be read as an upper bound of roughly two points, of the same order as run-to-run variation. The ANLI gap (9 to 14 points) is an order of magnitude larger and is unaffected.

\paragraph{The contribution of $K$ is measured on SNLI only.} We did not run the no-$K$ ablation on ANLI, so the usefulness of multi-hop \texttt{ConceptNet} retrieval for adversarial inference remains open. The SNLI result does not settle it, and arguably says little about it: SNLI premises and hypotheses share most of their vocabulary, so $\mathcal{P} \setminus \mathcal{H}$ and $\mathcal{H} \setminus \mathcal{P}$ are small and there is little for a bridge to connect, whereas ANLI hypotheses are written to avoid lexical overlap, which is precisely the configuration the two-hop pass was designed for. In future work, we will assess whether $K$ helps there, or whether its 80 edges mostly add noise to an already truncated input.

\paragraph{What the vocabulary can and cannot express.} The inventory is closed by construction: the \texttt{enum} makes an out-of-vocabulary relation impossible, so every predicate of the source proposition is either mapped onto one of the 28 relations or lost. The normal form of Section~\ref{extract} absorbs the bulk of English predication --- argument structure, adjunction, possession, class membership, negation --- which is why the loss is almost invisible on SNLI. Three families of meaning, however, have no faithful target. First, temporal and spatial granularity collapses onto \texttt{HasContext} and \texttt{AtLocation}, so \textit{since 1953} and \textit{in 1953} become indistinguishable. Second, scalar and comparative predication (\textit{taller than}, \textit{most of}, \textit{at least three}) has no relation of its own, and cardinality survives only as a \texttt{HasProperty} object on an explicit numeral. Third, propositional attitudes and modality (\textit{X believes that Y}, \textit{X may Y}) are flattened, since no relation embeds a proposition under an operator. Section~\ref{sec:price} shows that the first two families, together with coreference errors and the truncation of long premises, account for most of the residual gap on ANLI, where they are frequent, and are rare in SNLI captions.

\begin{samepage}
\section*{Acknowledgments}

This work was supported by the French National Research Agency (ANR) through the EQUATION project (ANR-25-CE23-2916).
The authors acknowledge the computing resources provided by the Lab-IA cluster of the Saclay-IA platform within the Mésocentre Paris-Saclay at Université Paris-Saclay.

\end{samepage}

\bibliography{custom}

\appendix

\section{Extraction Prompt}
\label{prompt}

The prompt sent to the extractor is reproduced here in four parts. Figure~\ref{fig:prompt-relations} gives the task statement and the 28-relation vocabulary, Figure~\ref{fig:prompt-rules} the normal form imposed on nodes and on argument structure, Figure~\ref{fig:prompt-shots} the nine few-shot pairs, and Figure~\ref{fig:prompt-schema} the JSON schema passed to the constrained decoder.

\begin{figure*}[t]
\scriptsize
\begin{verbatim}
You convert ONE English sentence (an atomic proposition) into knowledge-graph triples using
  the ConceptNet vocabulary. Output a JSON array of triples [{"h": head, "r": relation, "t":
  tail}, ...] and nothing else.
RELATIONS -- use ONLY these ConceptNet relations:

- IsA: class or type membership: [dog, IsA, animal]
- InstanceOf: named entity is an instance of a class: [parma, InstanceOf, city]
- HasProperty: attribute, state or cardinality: [coat, HasProperty, red] ; [man, HasProperty, two]
- NotHasProperty: negated attribute: [room, NotHasProperty, empty]
- CapableOf: the subject performs the action: [person, CapableOf, jump]
- NotCapableOf: the subject does NOT perform the action: [politician, NotCapableOf, attend_meeting]
- ReceivesAction: the object undergoes the action: [ball, ReceivesAction, kick]
- AtLocation: spatial location: [man, AtLocation, park] ; [jump, AtLocation, street]
- LocatedNear: proximity: [bench, LocatedNear, fountain]
- HasA: possession or attachment: [person, HasA, horse]
- PartOf: part-whole: [wheel, PartOf, car]
- MadeOf: material: [knife, MadeOf, steel]
- UsedFor: purpose or instrument of an action: [knife, UsedFor, cut]
- Desires: wants / intends: [child, Desires, ice_cream]
- NotDesires: does not want: [cat, NotDesires, bath]
- CausesDesire: makes one want: [heat, CausesDesire, swim]
- Causes: causal link: [rain, Causes, wet]
- HasSubevent: sub-action happening within the action: [cook, HasSubevent, stir]
- HasPrerequisite: necessary precondition: [win, HasPrerequisite, compete]
- MotivatedByGoal: action done for a goal: [train, MotivatedByGoal, competition]
- MannerOf: specific way of doing (verb-verb): [sprint, MannerOf, run]
- HasContext: temporal or contextual setting: [sleep, HasContext, night]
- CreatedBy: creator or origin: [painting, CreatedBy, artist]
- DefinedAs: definition or equivalence: [sum, DefinedAs, total]
- Synonym: same meaning: [couch, Synonym, sofa]
- Antonym: opposite meaning: [hot, Antonym, cold]
- SimilarTo: similar: [jog, SimilarTo, run]
- DistinctFrom: mutually exclusive alternative: [indoors, DistinctFrom, outdoors]
\end{verbatim}
\caption{System prompt, part 1: task statement and the 28-relation ConceptNet extraction vocabulary.}
\label{fig:prompt-relations}
\end{figure*}

\begin{figure*}[t]
\scriptsize
\begin{verbatim}
NODE RULES:
- a node is a lowercase English lemma: the bare HEAD word, without its modifiers
- NEVER glue modifier+noun or verb+object into one node (wrong: old_man, go_package,
  wear_jacket); each modifier becomes its own HasProperty triple on the head noun
- multi-word nodes ONLY for proper names (new_york, forza_italia) and true lexical compounds
  (winter_hat, hot_dog); exception: non-intersective adjectives stay attached
  (former_senator, fake_gun)
- verbs: bare lemma (jump, not jumps/jumping); phrasal verbs keep their particle (jump_over)
- nouns keep their lexical form (linguistics stays linguistics); never use bare adverbs or
  prepositions as nodes (outside, just, very)
- numerals ONLY if the number is explicit: "two men" -> [man, HasProperty, two]; a bare
  plural ("the sisters") NEVER yields a numeral
- worn or printed identifiers are not quantities: "the number 2 jersey" -> [jersey,
  HasProperty, number_2]
STRUCTURE -- every content word of the sentence must land in at least one triple:
- agent of a verb: [agent, CapableOf, verb] -- the agent ALWAYS takes CapableOf, never
  HasProperty and never ReceivesAction
- direct object: [object, ReceivesAction, verb] -- for every transitive verb emit BOTH the
  agent triple AND the patient triple
- passive voice: the subject is the patient ("the book was written" -> [book,
  ReceivesAction, write]); a "by X" agent gets [X, CapableOf, verb]
- recipient or beneficiary ("to/for X"): [verb, HasContext, X] -- NEVER ReceivesAction
- destination ("walk to school"): [verb, MotivatedByGoal, school] -- not AtLocation
- place where it happens: [x, AtLocation, place]; time, date or period: [x, HasContext,
  time] -- dates are context, never HasProperty
- instrument: [instrument, UsedFor, verb] -- purpose: [verb, MotivatedByGoal, goal]
- possession: [owner, HasA, thing] -- part or aspect of: [part, PartOf, whole]
- role or profession: [person, IsA, role]; named entity class: [name, InstanceOf, class]
- created works: [work, CreatedBy, creator] -- the WORK is always the head, in active voice
  too: "Panini wrote a grammar" -> [grammar, CreatedBy, panini], NEVER [panini, CreatedBy,
  grammar]
- "outside/near X": attach the real place directly ([fight, LocatedNear, deli]), never a
  bare "outside" node
- NEGATION: use NotCapableOf / NotHasProperty / NotDesires, with the full verb phrase as
  tail when needed ([politician, NotCapableOf, attend_meeting]); NEVER create not_* nodes.
Produce 2 to 8 triples.
\end{verbatim}
\caption{System prompt, part 2: the normal form imposed on nodes and on argument structure.}
\label{fig:prompt-rules}
\end{figure*}

\begin{figure*}[t]
\scriptsize
\begin{verbatim}
IN : A person is training his horse for a competition.
OUT: [person, CapableOf, train] [horse, ReceivesAction, train] [person, HasA, horse] [train,
     MotivatedByGoal, competition]

IN : Two young girls are playing soccer in the park.
OUT: [girl, HasProperty, two] [girl, HasProperty, young] [girl, CapableOf, play] [soccer,
     ReceivesAction, play] [play, AtLocation, park]

IN : The chef is cutting fresh bread with a steel knife.
OUT: [chef, CapableOf, cut] [bread, ReceivesAction, cut] [bread, HasProperty, fresh] [knife,
     UsedFor, cut] [knife, MadeOf, steel]

IN : The politicians did not attend the meeting.
OUT: [politician, NotCapableOf, attend_meeting]

IN : The sisters are selling lemonade to a customer.
OUT: [sister, CapableOf, sell] [lemonade, ReceivesAction, sell] [sell, HasContext, customer]

IN : A boy wearing a jersey with the number 2 walks to school.
OUT: [boy, CapableOf, wear] [jersey, ReceivesAction, wear] [jersey, HasProperty, number_2]
     [boy, CapableOf, walk] [walk, MotivatedByGoal, school]

IN : The Indian grammarian Panini wrote the Ashtadhyayi, a description of Sanskrit, in the
  4th century BCE.
OUT: [panini, IsA, grammarian] [panini, HasProperty, indian] [panini, CapableOf, write]
     [ashtadhyayi, ReceivesAction, write] [ashtadhyayi, CreatedBy, panini] [ashtadhyayi,
     DefinedAs, description] [description, HasContext, sanskrit] [write, HasContext,
     4th_century_bce]

IN : The Parma trolleybus system has been in operation since 1953.
OUT: [parma_trolleybus_system, InstanceOf, trolleybus_system] [parma_trolleybus_system,
     AtLocation, parma] [parma_trolleybus_system, ReceivesAction, operate] [operate,
     HasContext, since_1953]

IN : A man sleeps on the couch because he is tired.
OUT: [man, CapableOf, sleep] [sleep, AtLocation, couch] [man, HasProperty, tired] [tired,
     Causes, sleep]
\end{verbatim}
\caption{The nine few-shot pairs, in compact triple notation. They are sent as alternating user and assistant turns, the assistant turns containing JSON arrays of \texttt{\{h, r, t\}} objects.}
\label{fig:prompt-shots}
\end{figure*}

\begin{figure*}[t]
\footnotesize
\begin{verbatim}
{
  "type": "array",
  "items": {
    "type": "object",
    "properties": {
      "h": {"type": "string", "maxLength": 60},
      "r": {"enum": [
        "IsA", "InstanceOf", "HasProperty", "NotHasProperty", "CapableOf",
        "NotCapableOf", "ReceivesAction", "AtLocation", "LocatedNear",
        "HasA", "PartOf", "MadeOf", "UsedFor", "Desires", "NotDesires",
        "CausesDesire", "Causes", "HasSubevent", "HasPrerequisite",
        "MotivatedByGoal", "MannerOf", "HasContext", "CreatedBy",
        "DefinedAs", "Synonym", "Antonym", "SimilarTo", "DistinctFrom"
      ]},
      "t": {"type": "string", "maxLength": 60}
    },
    "required": ["h", "r", "t"],
    "additionalProperties": false
  },
  "minItems": 1,
  "maxItems": 8
}
\end{verbatim}
\caption{JSON schema passed to the constrained decoder. The \texttt{enum} lists the 28 extraction relations of Figure~\ref{fig:prompt-relations}.}
\label{fig:prompt-schema}
\end{figure*}

\section{Reproduction Code}
\label{repro}

The code is available at \url{https://github.com/l-pommeret/graph-nli}. Commands assume the ConceptNet 5.7 English
assertions in \texttt{data/conceptnet/}. Sequential fine-tuning starts from the SNLI
checkpoint selected on the validation split. The atomic proposition datasets are
released on the HuggingFace Hub.\footnote{\url{https://huggingface.co/datasets/Younes2E/atomic-snli} and \url{https://huggingface.co/datasets/Younes2E/atomic-anli}}
Figure~\ref{fig:pipeline} gives the end-to-end commands.

\begin{figure*}[t]
\footnotesize
\begin{verbatim}
# 0. atomic propositions (released on the HuggingFace Hub)
huggingface-cli download --repo-type dataset Younes2E/atomic-snli --local-dir data/atomic/snli
huggingface-cli download --repo-type dataset Younes2E/atomic-anli --local-dir data/atomic/anli

# 1. triples (GPU, vLLM)
python -m nli.extract --inputs data/atomic/snli/{train,validation,test}.jsonl \
    --out out/snli/triples.jsonl

# 2. three-graph datasets (P, H, K)
python -m nli.dataset --atomic data/atomic/snli/train.jsonl \
    --triples out/snli/triples*.jsonl --out out/snli/train.dataset.jsonl

# 3. fine-tuning on SNLI, then on ANLI from the SNLI checkpoint
python -m nli.finetune --train out/snli/train.dataset.jsonl \
    --val out/snli/validation.dataset.jsonl --test out/snli/test.dataset.jsonl \
    --out out/models/snli-s0 --seed 0 --no-grad-ckpt --epochs 2 --max-len 1024
python -m nli.finetune --model out/models/snli-s0/checkpoint-8000 \
    --train out/anli/train_r{1,2,3}.dataset.jsonl \
    --val out/anli/dev_r{1,2,3}.dataset.jsonl \
    --test out/anli/test_r{1,2,3}.dataset.jsonl out/snli/test.dataset.jsonl \
    --out out/models/anli-phase2 --max-len 1536 --epochs 3
\end{verbatim}
\caption{End-to-end pipeline, from atomic propositions to the fine-tuned classifier.}
\label{fig:pipeline}
\end{figure*}

\end{document}